\documentclass[10pt,twocolumn,letterpaper]{article}

\usepackage[pagenumbers]{cvpr} 

\usepackage{graphicx}
\usepackage{amsmath}
\usepackage{amssymb}
\usepackage{booktabs}
\usepackage[dvipsnames]{xcolor}
\usepackage{multirow}
\usepackage{rotating}
\usepackage{color}
\usepackage{pifont}
\usepackage{makecell}

\newcommand{\cmark}{\ding{51}\xspace}%
\newcommand{\cmarkg}{\textcolor{lightgray}{\ding{51}}\xspace}%
\newcommand{\xmark}{\ding{55}\xspace}%
\newcommand{\xmarkg}{\textcolor{lightgray}{\ding{55}}\xspace}%
\newcommand{\three}{3}
\newcommand{\threeg}{\textcolor{lightgray}{3}}
\newcommand{\five}{5}
\newcommand{\fiveg}{\textcolor{lightgray}{5}}

\usepackage[pagebackref,breaklinks,colorlinks]{hyperref}

\usepackage[capitalize]{cleveref}
\crefname{section}{Sec.}{Secs.}
\Crefname{section}{Section}{Sections}
\Crefname{table}{Table}{Tables}
\crefname{table}{Tab.}{Tabs.}

\def\cvprPaperID{3933} 
\def\confName{CVPR}
\def\confYear{2022}

\begin{document}

\title{SemAffiNet: Semantic-Affine Transformation for Point Cloud Segmentation}

\author{Ziyi Wang\textsuperscript{1,2}, Yongming Rao\textsuperscript{1,2}, Xumin Yu\textsuperscript{1,2}, Jie Zhou\textsuperscript{1,2}, Jiwen Lu\textsuperscript{1,2}\thanks{Corresponding author}\\
\textsuperscript{1}Department of Automation, Tsinghua University\\
\textsuperscript{2}Beijing National Research Center for Information Science and Technology \\
{\tt\small wziyi20@mails.tsinghua.edu.cn; raoyongming95@gmail.com; } \\
{\tt \small yuxm20@mails.tsinghua.edu.cn; \{jzhou, lujiwen\}@tsinghua.edu.cn} \\
}
\maketitle

\begin{abstract}

Conventional point cloud semantic segmentation methods usually employ an encoder-decoder architecture, where mid-level features are locally aggregated to extract geometric information. However, the over-reliance on these class-agnostic local geometric representations may raise confusion between local parts from different categories that are similar in appearance or spatially adjacent. To address this issue, we argue that mid-level features can be further enhanced with semantic information, and propose \textbf{semantic-affine transformation} that transforms features of mid-level points belonging to different categories with class-specific affine parameters. Based on this technique, we propose \textbf{SemAffiNet} for point cloud semantic segmentation, which utilizes the attention mechanism in the Transformer module to implicitly and explicitly capture global structural knowledge within local parts for overall comprehension of each category. We conduct extensive experiments on the ScanNetV2 and NYUv2 datasets, and evaluate semantic-affine transformation on various 3D point cloud and 2D image segmentation baselines, where both qualitative and quantitative results demonstrate the superiority and generalization ability of our proposed approach. Code is available at \url{https://github.com/wangzy22/SemAffiNet}.

\end{abstract}

\begin{figure}
	\centering
	\includegraphics[width=0.95\linewidth]{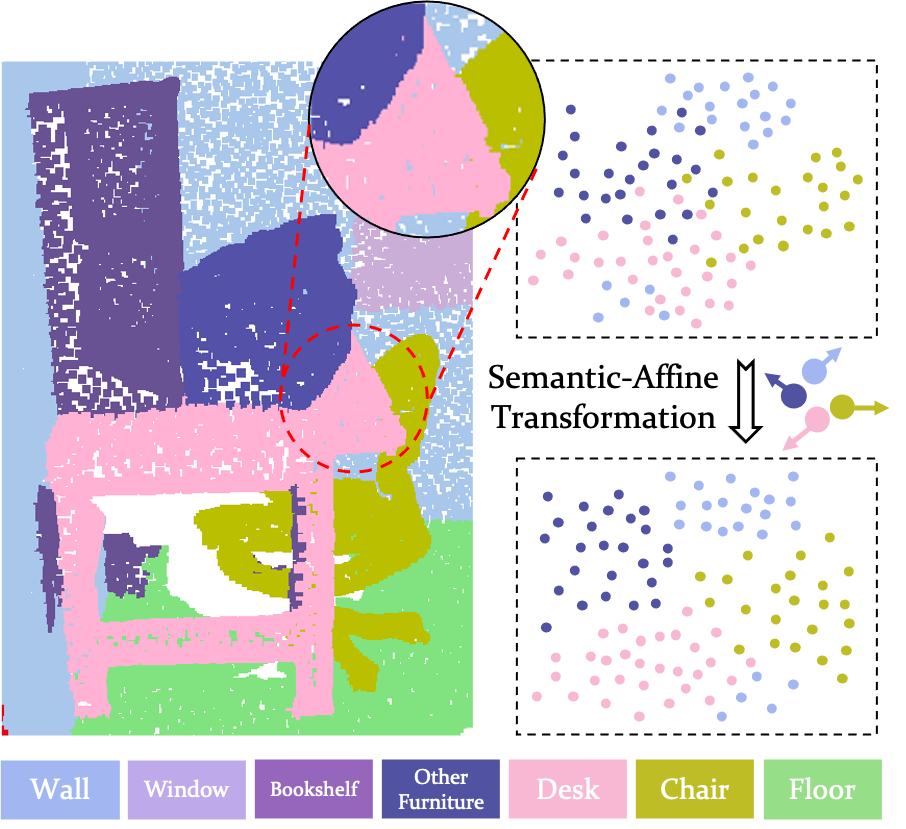}
	\caption{Illustration of Semantic-Affine Transformation. The left figure is the input point cloud, with different colors referring to different categories. We select one local part (red circle) and extract its mid-level features, resulting in the top-right figure. Some representations of the points from different categories are entangled with each other in the embedding space, which may be caused by the appearance similarities or spatial adjacency. We propose to perform the semantic-affine transformation on these mid-level features, predicting particular affine parameters for each category respectively. Therefore, once obtaining classification predictions of mid-level points, we can pull points from the same category closer while pushing points from different categories apart via semantic-affine transformation, as is shown in the bottom-right figure.}
	\label{fig:overview}
	\vspace{-3mm}
\end{figure}

\section{Introduction}
\label{sec:intro}

Point cloud semantic segmentation is a fundamental task for both structural representation learning~\cite{qi2017pointnet, choy20194d, wang2019dynamic} and stereoscopic scene understanding\cite{rethage2018fully, dai20183dmv, huang2019texturenet} in computer vision. It aims at partitioning the scene space into semantic-meaningful regions based on conformation and geometry knowledge inherited in point cloud layouts. Its successful applications in autonomous driving, robotic manipulation, and virtual reality have been motivating researchers to develop more fine-grained and more accurate solutions. 

Recent methods for point cloud segmentation usually adopt an encoder-decoder architecture as image semantic segmentation~\cite{zhao2017pyramid, ronneberger2015u, badrinarayanan2017segnet, chen2017deeplab, chen2017rethinking}, ranging from voxel based-ones~\cite{choy20194d, graham20183d, su2018splatnet, zhang2020polarnet, zhou2020cylinder3d} to point-based ones~\cite{qi2017pointnet++, li2018pointcnn, wu2019pointconv, thomas2019kpconv}. 
Despite the popularity of the encoder-decoder architecture, there still exists the \textit{local confusion problem} as shown in Figure~\ref{fig:overview}. On the one hand, there are local parts from different categories but with similar shapes, such as the similar legs of chairs and desks. On the other hand, adjacent local parts are blended in the input space and may obfuscate the model during segmentation, leaving ambiguous segment boundaries. The reasons are two folds: the heavy use of the local aggregation during feature processing, and the class-agnostic nature of the mid-level features. In the commonly-used encoder-decoder architecture, the mid-level features of the decoder are locally aggregated via convolution or set abstraction. The limitation of receptive fields produces similar feature vectors for visually-similar local parts, and the aggregation operation results in entangled mid-level features for spatially-adjacent local parts. Therefore, it is insufficient to use geometric-only information and the encoder-decoder architecture demands more knowledge to separate similar and entangled local representations. One possible solution to this problem is alleviating the reliance on geometric knowledge and introducing additional semantic information to enrich mid-level features. 
However, most existing literature fails to fully exploit semantic knowledge in the network design of the encoder-decoder architecture, as semantic annotations are mostly used for data augmentation~\cite{nekrasov2021mix3d,chen2020pointmixup,zhang2021pointcutmix} or supervision on final prediction~\cite{qi2017pointnet,qi2017pointnet++,wang2019dynamic}. Therefore, the mid-level features from the intermediate layers are only implicitly or weakly supervised via gradient descent, making them almost class-agnostic.

To address the \textit{local confusion problem}, we propose \textbf{Semantic-Affine Transformation} to transform mid-level decoder features with class-specific affine parameters that encode semantic information, which explicitly pulls features from the same category closer and pushes features from different categories apart. 
In this way, we enhance the semantic representation ability of mid-level features and boost semantic segmentation performance.
Based on the proposed semantic-affine transformation, we design a semantic-aware network named \textbf{SemAffiNet} and introduce Transformer~\cite{vaswani2017attention} to manage semantic information both implicitly and explicitly. The Transformer encoder implicitly communicates geometric information across modalities via the self-attention technique, while the special design of class queries in the Transformer decoder performs explicit semantic-aware reasoning to predict semantic-affine parameters via the cross-attention mechanism.  We conduct extensive experiments on the ScanNetV2~\cite{dai2017scannet} dataset and outperform the previous state-of-the-art BPNet~\cite{hu2021bidirectional} baselines. We also evaluate on the NYUv2~\cite{NYU2012Silberman} dataset to verify the generalization ability of the SemAffiNet model. As the core of SemAffiNet, the proposed semantic-affine transformation is evaluated on both 3D point cloud and 2D image segmentation baselines under various settings, revealing the generalization ability of the proposed transformation.

In conclusion, the contributions of our paper can be summarized as follows: (1) We propose Semantic-Affine Transformation to enhance the semantic representation ability of mid-level features in encoder-decoder segmentation architecture. (2) We propose SemAffiNet to perform semantic-aware segmentation both explicitly and implicitly via special designs of Transformer modules. (3) We conduct experiments on various datasets under different settings, revealing the superiority and generalization ability of our method.

\section{Related Work}

\noindent \textbf{Point Cloud Semantic Segmentation.} Existing point cloud semantic segmentation methods can be divided into four categories: voxel-based, point-based, projection-based, and hybrid models. 
Voxel-based methods aim at partitioning 3D space into ordered voxels and translating 2D convolutional encoder-decoder architectures to 3D conditions, leading by VoxNet~\cite{maturana2015voxnet}. The heavy time expense and memory cost have been addressed by later researches, including sparse convolution~\cite{graham20183d, choy20194d}, efficient data structure migration~\cite{riegler2017octnet, klokov2017escape} and novel voxelization techniques~\cite{zhang2020polarnet, su2018splatnet, zhou2020cylinder3d}.
Point-based methods directly process points and aggregate local information instead of using conventional regular convolution kernel, leading by PointNet~\cite{qi2017pointnet} and PointNet++~\cite{qi2017pointnet++}. Nowadays point-based methods has become the mainstream of point cloud cognition tasks and has been developed into many branches, including MLP-based~\cite{qi2017pointnet, qi2017pointnet++, fan2021scf, deng2021ga, hu2020randla}, convolution-based~\cite{wu2019pointconv, li2018pointcnn, thomas2019kpconv, xu2021paconv} and graph-based~\cite{wang2019dynamic, li2018adaptive} posterity. 
Projection-based methods are mostly designed for efficient processing, including image projection~\cite{aksoy2020salsanet, cortinhal2020salsanext} and spherical projection~\cite{wu2018squeezeseg, wu2019squeezesegv2, milioto2019rangenet++, alonso20203d}. 
Hybrid methods are more complicated systems that combine different processing methods or fuse different modality information. Some approaches combine long-range correlations from voxel-based methods and meticulous details from point-based methods~\cite{liu2019point, ye2021drinet, tang2020searching, cheng20212}, while other approaches fuse 2D and 3D knowledge~\cite{quan2016fusionnet, hu2021bidirectional}.

\begin{figure*}[tb]
	\centering
	\includegraphics[width=1\linewidth]{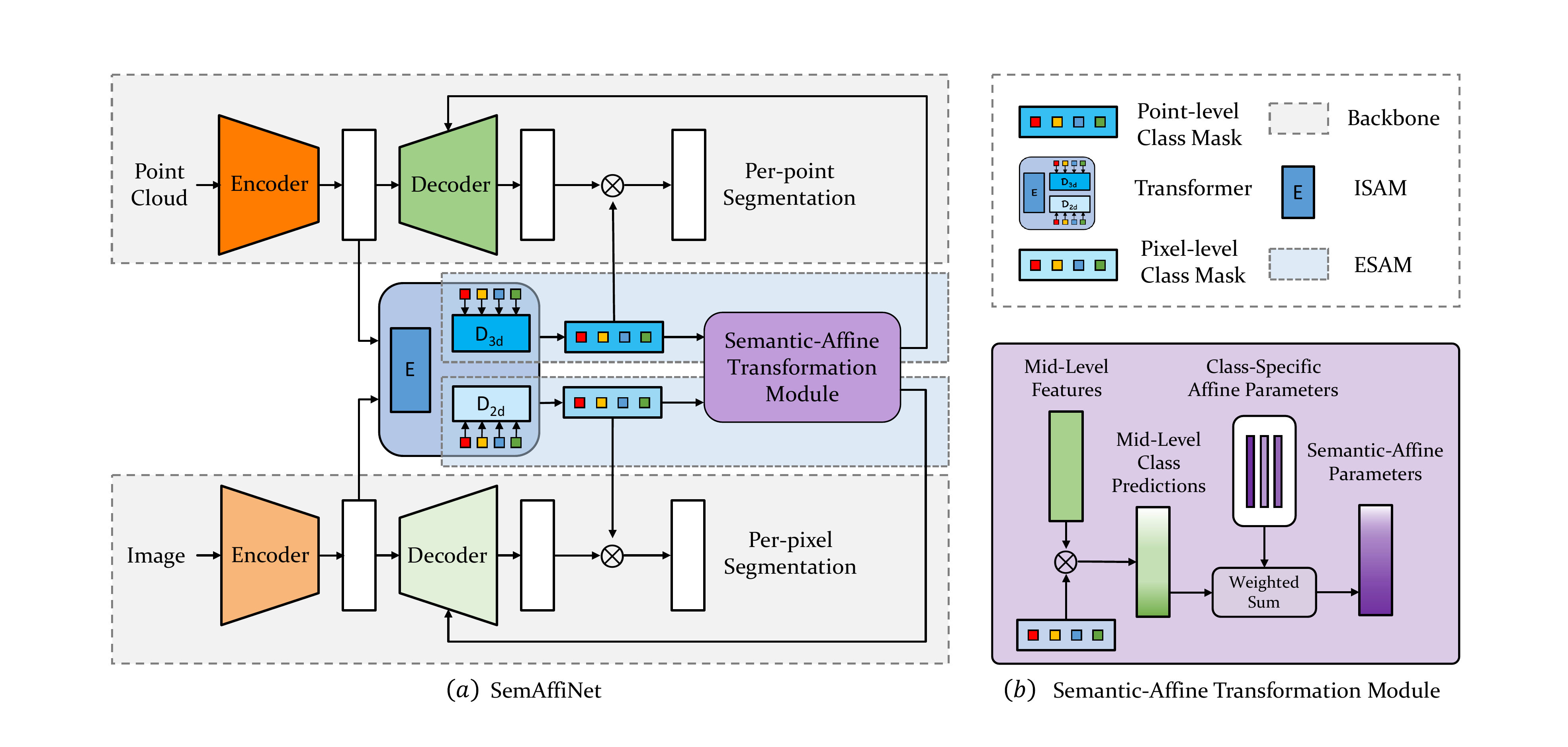}
	\caption{Illustration of the proposed network architecture. $(a)$ shows the pipeline of the SemAffiNet, which consists of two backbone branches (gray), an implicit semantic-aware module (ISAM, dark blue), and two explicit semantic-aware modules (ESAM, light blue dash square). Moreover, ESAM is composed of a Transformer decoder and a Semantic-Affine Transformation module, which is further illustrated in $(b)$. We calculate the weighted sum of class-specific affine parameters to obtain semantic-affine parameters for mid-level points, with mid-level classification confidences as linear combination weights.}
	\label{fig:arch}
	\vspace{-3mm}
\end{figure*}

\vspace{4pt}

\noindent \textbf{Semantic-Aware Segmentation.}
As semantic information is critical to the segmentation task, some works elaborately design special semantic-aware approaches to boost semantic segmentation performance~\cite{qiu2021semantic, lu2021cga, gong2021omni}.  
Some methods aim at reasoning context relations differently between same-category pairs and distinct-categories pairs. DependencyNet~\cite{liu2021exploit} for image segmentation unifies dependency reasoning at three semantic levels: intra-class, inter-class and global. CGANet~\cite{lu2021cga} for point cloud segmentation utilizes different aggregation strategies between the same category and different categories. 
Other methods propose multi-scale supervision to realize comprehensive semantic guidance. In 2D vision, CPM~\cite{wei2016convolutional} introduces intermediate supervision periodically while MSS-net~\cite{ke2018multi} proposes layer-wise loss. In 3D vision, RFCC~\cite{gong2021omni} puts forward omni-supervision on all levels of the decoder layers. 

Different from the above methods, our method achieves comprehensive semantic awareness via semantic-affine transformation for mid-level features. 
Therefore, we don't need different aggregation modules that increase model scale. 
Additionally, our semantic guidance for intermediate layers is stronger than merely multi-level supervisions.

\vspace{4pt}

\noindent \textbf{Transformer for Segmentation.} Transformer~\cite{vaswani2017attention} has achieved a great success in many computer vision tasks, such as classification\cite{dosovitskiy2020image, liu2021swin}, detection\cite{carion2020end}, and reconstruction\cite{yu2021pointr}. Recent works employ the attention mechanism in Transformer to exploit long-range correlations for deeper context comprehension and better segmentation results\cite{zheng2021rethinking, zhao2021point, guo2021pct}. Maskformer\cite{cheng2021maskformer} proposes a mask classification model that utilizes Transformer to predict binary masks and unifies both semantic- and instance-level segmentation. SOTR\cite{guo2021sotr} proposes to dynamically generating instance segmentation masks based on Transformer attention module.

While we utilize a similar mask classification structure as Maskformer in our SemAffiNet, we migrate this idea from 2D image processing to 3D point cloud understanding, which is not trivial. Moreover, we propose semantic-affine transformation to further enhance the mask classification pipeline, which brings more progression than mask classification according to our ablation studies.

\section{Approach}
In the following section, we will first give an overview of the proposed SemAffiNet in Section~\ref{sec:semaffinet}. Then we will present details of the architecture, introducing the proposed semantic affine transformation in Section~\ref{sec:sat}, revealing how we wrap it into a plug-and-play explicit semantic-aware module in Section~\ref{sec:esam}, and presenting the auxiliary implicit semantic-aware module in Section~\ref{sec:isam}. Finally, we will introduce the loss function design in Section~\ref{sec:loss}.

\subsection{Overview}
\label{sec:semaffinet}

We elaborately design SemAffiNet to perform the semantic-affine transformation on mid-level features from the conventional encoder-decoder model. Figure~\ref{fig:arch} shows the overall architecture, which can be divided into the following three parts: (1) Backbone, (2) the Explicit Semantic-Aware Module (ESAM), (3) the Implicit Semantic-Aware Module (ISAM).

First of all, the backbone choice of SemAffiNet is flexible and our proposed modules can be easily added to any encoder-decoder segmentation architecture. We choose BPNet~\cite{hu2021bidirectional} that consists of two encoder-decoder branches for 2D and 3D modalities to evaluate our proposed semantic-affine transformation. Please refer to the BPNet paper or our supplementary material for further details.

Most importantly, ESAM wraps our proposed semantic-affine transformation into a plug-and-play module, that explicitly exploits semantic information in mid-level features with specially designed Transformer decoders. As shown Figure~\ref{fig:arch}, we employ two ESAM blocks in light blue dash square to manage semantic knowledge from different domains respectively. The Transformer decoder utilizes the cross-attention mechanism to obtain long-range dependencies for better semantic perception, while the following Semantic-Aware Transformation Module transforms mid-level features of backbone decoders with class-specific affine parameters to enlarge semantic distinctions across categories.

Last but not least, ISAM utilizes the self-attention mechanism in the Transformer encoder to enhance high-level features that are outputs from the backbone encoder and inputs to ESAM. The proposed ISAM fuses multi-modality information and realizes implicit semantic awareness. 

\begin{figure}
	\centering
	\includegraphics[width=1\linewidth]{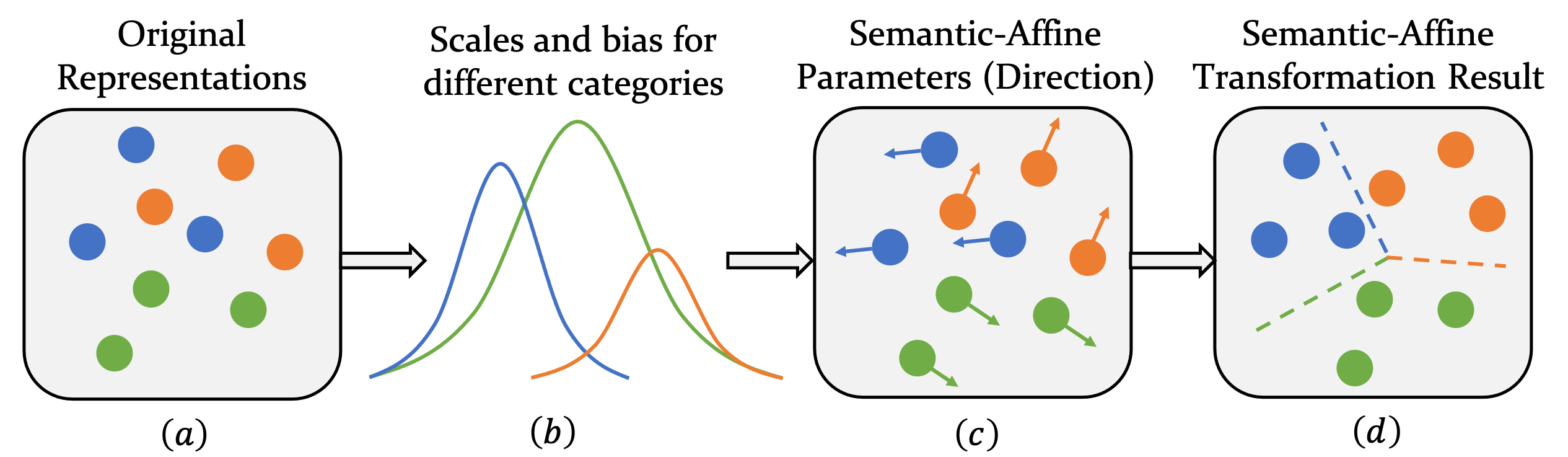}
	\caption{Illustrations of the Semantic-Affine Transformation. (a) is the original representations of mid-level points. (b) is the underlying distinctive scales and bias for different categories. The arrow directions in (c) represent the regressed class-specific affine parameters. (d) shows the semantic-affine transformation results of mid-level features, which are more classification-friendly.}
	\label{fig:sat}
	\vspace{-3mm}
\end{figure}

\subsection{Semantic-Affine Transformation}
\label{sec:sat}
As semantic-affine transformation is the key contribution of our paper, we will first introduce its conceptual and technical details in this sub-section. The core idea is predicting semantic-affine parameters for each category and then determining affine parameters for each point based on its classification prediction.

Suppose there are $N$ classes and the shape of mid-level features $\mathbf{f}^i$ at layer $i$ is $(n_i, d_i)$, indicating that there are $n_i$ points at layer $i$ and each point is represented by a $d_i$-dim feature vector. From Section~\ref{sec:esam} we can predict categories for each point $p^i_j$ at layer $i$ and obtain a classification confidence vector $\mathbf{a}^i_j=[a^i_{j1}, a^i_{j2}, \cdots, a^i_{jN}]$, $0 \leq a^i_{jk} \leq 1, 1 \leq j \leq n_i, 1 \leq k \leq N, \sum_{k}a^i_{jk} = 1$, where $a^i_{jk}$ denotes confidence of point $p_j$ at layer $i$ belonging to class $k$. Simultaneously, we regress semantic-affine parameters for each class at layer $i$, including scale factor $\mathbf{s}^i=\{s^i_k \in \mathbb{R}^{d_i},1 \leq k \leq N, s^i_{kl} \ge 0, 1 \leq l \leq d_i\}$ and offset bias $\mathbf{b}^i=\{b^i_k \in \mathbb{R}^{d_i},1 \leq k \leq N\}$. Further technical details about the learning process of these semantic-aware parameters can be found in Section~\ref{sec:esam}.

Then we can obtain the semantic-specific affine parameters $S^i_j, B^i_j$ for each point $p_j$ at layer $i$ via linear combination of the affine parameters family $\mathbf{s}^i, \mathbf{b}^i$ based on per-point classification confidence vector $\mathbf{a}^i_j$:
\begin{equation}
    S^i_j = \sum_{k}{a^i_{jk}s^i_k}, \quad B^i_j = \sum_{k}{a^i_{jk}b^i_k}
\label{eq:semparam}
\end{equation}

Once normalized with zero-mean and unit covariance, $\hat{f}^i_{j}=(f^i_{j} - \mu(f^i_j)) / \sigma(f^i_j)$ can be further enhanced with the semantic-affine transformation to obtain semantic-aware mid-level feature $\Tilde{f}^i_{j}$, which is used to replace $f^i_j$:
\begin{equation}
    \Tilde{f}^i_{j} = S^i_j \hat{f}^i_{j} + B^i_j
\label{eq:semaffine}
\end{equation}

Note that we implement a soft semantic-affine parameters assignment that introduces a linear combination instead of a hard one that restricts the searching space within the exact value of $\textbf{s}, \textbf{b}$. In other words, the hard assignment strategy only considers the highest confidence scores of the category predictions and chooses the exact affine parameters accordingly. The reason is that the mid-level points are aggregation results of itself and its neighbors in the adjacent lower level, and the neighbors may have different categories from the center query point. Therefore, the mid-level points $p_i$ may represent multiple classes when its corresponding patch in lower levels lies at the edges. We will further discuss this issue in Section~\ref{sec:loss}. 

We illustrate the principle of the semantic-affine transformation in Figure~\ref{fig:sat}. The original representations of points from different categories in $(a)$ are entangled with each other. Then we train a network to capture the underlying distinctions of scales and bias between different categories, as is shown in $(b)$. Then we express these distinctions with semantic-affine parameters, which are demonstrated as arrow directions in $(c)$. Finally, in $(d)$, the semantic-affine transformation explicitly transforms mid-level features with similar category distributions towards similar scales and offsets, thus pulling them closer. On the contrary, for mid-level features with distinctive category distributions, their discrepancy in scales and offsets pushes them further apart.

According to the discussions above, the most important prerequisites for semantic-affine transformation are two-fold. The first one is a precise class predictor, predicting accurate category distributions for mid-level points. The second one is a powerful semantic-aware module, regressing representative affine transformation parameters for each class. In Section~\ref{sec:esam}, we show that these two prerequisites can be satisfied by a multi-layer Transformer decoder.

\subsection{The Explicit Semantic-Aware Module}
\label{sec:esam}
Our goal is to wrap the learnable parameters of semantic-aware transformation introduced above into a plug-and-play module that can be implemented in most encoder-decoder-style semantic segmentation architecture. We propose an Explicit Semantic-Aware Module (ESAM) as a multi-layer Transformer decoder module to jointly and explicitly inference \textit{semantic class masks} and \textit{semantic affine parameters}.

\vspace{4pt}

\noindent \textbf{Cross attention in Transformer decoder.} The input to ESAM is the high-level feature $\mathbf{f}^{0}$, which is the output from the backbone encoder. Particularly, we design $N$ learnable class queries $q^{(c)}$ to enquire semantic-specific knowledge. Then each layer of ESAM utilizes the attention mechanism to reason semantic information from $\mathbf{f}^{0}$:
\begin{equation}
    \textrm{Attention}(Q,K,V) = \textrm{softmax}\left(\frac{QK^T}{\sqrt{d_k}}\right)V
\label{eq:attn}
\end{equation}
where $d_k$ is the scaling factor, and $Q,K,V$ are queries, keys, and values matrix. We employ class queries $q^{(c)}$ as $Q$, while $K, V$ are mapped embeddings of $\mathbf{f}^{0}$. From each layer of ESAM, we can obtain class-specific features $h \in \mathbb{R}^{N\times d_h}$ of $d_h$ dimensions that encode both class-wise semantic knowledge from $q^{(c)}$ and geometry knowledge of the scene from $\mathbf{f}^{0}$.

\vspace{4pt}

\noindent \textbf{Semantic-affine parameters.} Based on the multi-layer Transformer decoder architecture of ESAM, we utilize the intermediate outputs from ESAM mid-level layers to calculate the semantic-affine parameters for mid-level layers of the backbone decoder: $\mathbf{s}^{i}_k = \textrm{MLP}(h^{u}_k), \mathbf{b}^{i}_k = \textrm{MLP}(h^{u}_k)$,
where $1 \leq k \leq N$ denotes class $k$, and there is a one-to-one and order-preserved mapping between backbone decoder mid-layer $i$ and ESAM mid-layer $u$. The principle is using output features of deeper ESAM layer $u$ to calculate semantic-affine parameters of deeper backbone decoder layer $i$. Please refer to the supplementary material for detailed correspondence between $i$ and $u$. The outcoming scale and bias parameters $\mathbf{s}, \mathbf{b}$ are then used to perform semantic-affine transformation introduced in Section~\ref{sec:sat}.

\vspace{4pt}

\noindent \textbf{Semantic class masks.} The segmentation masks $M=\{m_k \in \mathbb{R}^{d_m}, 1 \leq k \leq N\}$ for $N$ classes per-point classification are calculated via the output $h^{-1}$ from the final layer of ESAM: $m_k = \textrm{MLP}(h^{-1}_k)$.
Then we implement dot product on class mask $m_k$ and per-point feature $\mathbf{f}^{i} = \{f^{i}_j\in \mathbb{R}^{d_i}, 1 \leq j \leq n^{i}\}$ to calculate the confidence matrix $A^{i} = \{a^{i}_{jk}\}$: $a^{i}_{jk} = m_k \cdot f^{i}_j$, where $a^{i}_{jk}$ indicates the confidence of point $p_j$ at layer $i$ of the backbone decoder belonging to class $k$ and is utilized in Section~\ref{sec:sat}.

Note that in conventional per-point segmentation methods such as FCN~\cite{long2015fully} or encoder-decoder-style architecture UNet~\cite{ronneberger2015u}, per-point feature $\mathbf{f}^{-1}$ from the final layer of the decoder is further processed by an MLP block as segmentation head to obtain per-point class predictions. However, instead of linearly combining channel-wise values via fully connected layers to predict class confidence, we implement the above mask prediction and dot production with per-point features to classify mid-level and final point features. The advantages of predicting class masks are two folds. On the one hand, the class masks have clearer interpretive meanings. Each mask $m_k$ represents the comprehensive feature of the class $k$, and the dot product between $m_k$ and the point feature measures the similarity between the point and the class. Therefore, the point is categorized into its most similar class. On the other hand, the class masks are more flexible to be implemented to predict categories of mid-level points. Once $M$ is obtained, the dot-product operation is more lightweight than the MLP forward calculation. Therefore, class mask prediction is more suitable to combine with our proposed semantic-affine transformation.

In conclusion, ESAM explicitly predicts semantic class masks and semantic-affine parameters via a multi-layer Transformer decoder with learnable class queries. Each class query represents a category and inquires class-specific information in scene geometric representations. Then the semantic class masks are used to perform more flexible and lightweight multi-level per-point classification, while semantic-affine parameters are applied to transform mid-level features of the backbone decoder.

\begin{figure}
	\centering
	\includegraphics[width=1\linewidth]{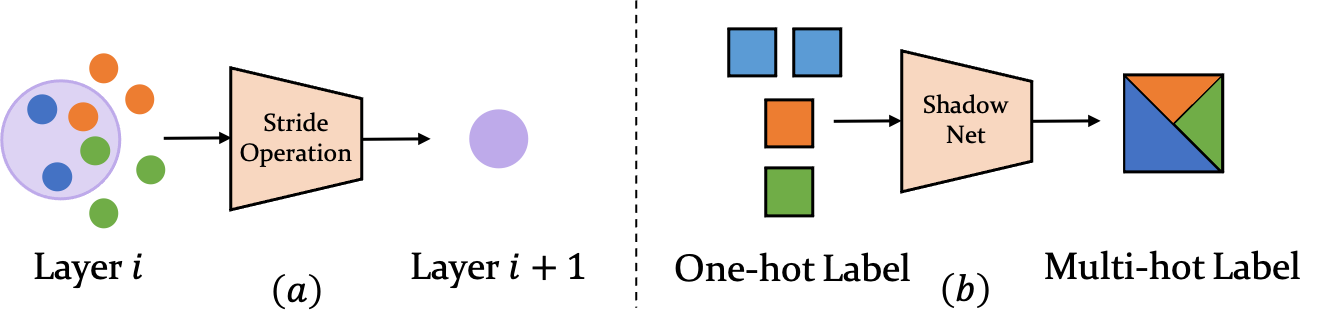}
	\caption{Illustration of ShadowNet that is designed to trace multi-hot ground truth label for mid-level points. Column $(a)$ shows the local aggregation of stride operation in the backbone encoder. Column $(b)$ shows how we record the class distribution of the points at the higher layer according to the class distribution of its corresponding patch at the lower layer.}
	\label{fig:shadow}
	\vspace{-3mm}
\end{figure}

\subsection{The Implicit Semantic-Aware Module}
\label{sec:isam}
Besides ESAM that explicitly reasons semantic information via specially designed learnable class queries in the Transformer decoder, we also design ISAM to implicitly exploit and fuse semantic knowledge from multi-modalities.

In our implementation, 2D high-level feature $\mathbf{f}^{0,2d}$ and 3D high-level feature $\mathbf{f}^{0,3d}$ are concatenated together to form mixed Transformer encoder input $\mathbf{f}^{0,mix}$. Then queries, keys, and values matrix are obtained via three different linear transformations of the input mix feature $\mathbf{f}^{0,mix}$. Therefore, the self-attention calculation in Equation~(\ref{eq:attn}) are performed at both inter-modality and intra-modality to obtain more representative high-level features $\bar{\mathbf{f}}^{0,2d}$ and $\bar{\mathbf{f}}^{0,3d}$. On the one hand, the intra-modality self-attention reasons long-range dependencies between local parts from the same modality, attaching global information to local part features. On the other hand, the inter-modality self-attention captures similarities between parts from different modalities, creates soft correspondences, and merges knowledge from the other modal to local features. Therefore, the output features of ISAM acquire both long-range global knowledge and multi-modal information, making them more robust and more representative. Thus the \textit{implicit} semantic-awareness is realized via intra-domain and inter-domain self-attention, where queries that are more semantic-similar to keys contribute more to updating the values matrix.   

\subsection{The Loss Function}
\label{sec:loss}
Following conventional supervised segmentation approaches, we use the Cross Entropy loss for vanilla 2D per-pixel segmentation and 3D per-point segmentation.

Additionally, since we predict category labels for mid-level points in backbone decoders, we calculate the Binary Cross Entropy loss for mid-level segmentation. In order to obtain the mid-level ground truth of backbone decoders, we design ShadowNets to trace the \textit{stride} operation in their corresponding backbone encoders. As is shown in Figure~\ref{fig:shadow}, stride operation in the encoder aggregates points $\{p^i_{j}\}$ within a local patch $P^i_{j'}$ at layer $i$ into a meta-point $p^{i+1}_{j'}$ at higher layer $i+1$. Suppose the one-hot label for point $p^i_{j}$ is $l^i_{j}$, then our ShadowNet assigns the meta-point with multi-hot label $l^{i+1}_{j'}$ that records labels of all points within its corresponding patch $P^i_{j'}$ at the lower layer $i$:
\begin{equation}
    l^{i+1}_{j'} = \min(1, \sum_{p^{i}_{j}\in P^{i}_{j'}} l^{i}_{j})
\end{equation}

In this way, the multi-hot ground truth labels represent class distributions of points at mid-level layers.

\section{Experiments}
In this section, we conduct extensive experiments on various datasets to verify the superiority of the proposed semantic-affine transformation and the SemAffiNet architecture, calculating the class-wise mean intersection over union (mIoU) as evaluation metrics. In Section~\ref{sec:main_result}, we will present the quantitative and qualitative results of the SemAffiNet, comparing them with previous works. Then in Section~\ref{sec:sat_analysis}, we will implement the semantic-affine transformation on both 3D point cloud and 2D image segmentation baselines under 
different conditions to prove its generalization ability. Furthermore, in Section~\ref{sec:ablation}, we will provide ablation studies to demonstrate the effectiveness of each proposed module. Finally, in Section~\ref{sec:limit}, we will discuss the limitations of our proposed approach. 
Additionally, experiments setup including datasets introduction and implementation details can be found in supplementary materials.

\begin{table}
  \centering
  \caption{Quantitative results on the ScanNetV2 dataset.  We compare both 3D and 2D mIoU with our baseline method BPNet. We also compare 3D mIoU with other works that use point cloud as input. Methods marked with * use additional 2D image input.}
  \setlength{\tabcolsep}{4mm}
  \begin{tabular}{@{}lcc@{}}
    \toprule
        Method              &   3D mIoU(\%)     &   2D mIoU(\%) \\
    \midrule
        PointNet++\cite{qi2017pointnet++}  &   53.5            &    -- \\
        PointConv\cite{wu2019pointconv}     &   61.0            &   -- \\
        PointASNL\cite{yan2020pointasnl}    &   63.5            &   -- \\
        MVPNet*\cite{jaritz2019multi}        &   66.4            &   -- \\
        KPConv\cite{thomas2019kpconv}       &   69.2            &   -- \\
        SparseConvNet\cite{graham20183d}    &   69.3            &   -- \\
        RFCR\cite{gong2021omni}             &   70.2            &   -- \\       
        FAConv*\cite{zhang2020fusion}        &   72.0            &   -- \\
        MinkowskiNet\cite{choy20194d}       &   72.2            &   -- \\ 
        Mix3D\cite{nekrasov2021mix3d}       &   73.6            &   -- \\
    \midrule
        BPNet*\cite{hu2021bidirectional} (5cm)& 70.6            &   65.1 \\
        SemAffiNet* (5cm)    &   \textbf{72.1}   &   \textbf{68.2} \\
    \midrule
        BPNet* (2cm)         &   72.5            &   72.7 \\
        SemAffiNet* (2cm)    &   \textbf{74.5}   &   \textbf{74.2} \\
    \bottomrule
  \end{tabular}
  \label{tab:san}
\end{table}

\begin{table}
  \centering
  \caption{NYUv2 2D image segmentation results (13-class task). We compare our SemAffiNet with typical RGB-D based methods and joint 2D-3D methods on dense pixel classification accuracy metric. Baseline results are from the BPNet~\cite{hu2021bidirectional} paper and the results of~\cite{dai2017scannet} is on the 11-class task.}
  \setlength{\tabcolsep}{5mm}
  \begin{tabular}{@{}cc@{}}
    \toprule
        Method                              &   Accuracy(\%) \quad \\
    \midrule
        SceneNet~\cite{handa2016understanding}                  &   52.5 \\
        Hermans \textit{et al.}~\cite{hermans2014dense}         &   54.3 \\
        SemanticFusion~\cite{mccormac2017semanticfusion}        &   59.2 \\
        ScanNet~\cite{dai2017scannet}                           &   60.7 \\
        3DMV~\cite{dai20183dmv}             &   71.2 \\
        BPNet~\cite{hu2021bidirectional}    &   73.5 \\
        SemAffiNet (Ours)  &   \textbf{78.3} \\
    \bottomrule
  \end{tabular}
  \label{tab:nyu}
\end{table}

\begin{figure*}[tb]
	\centering
	\includegraphics[width=1\linewidth]{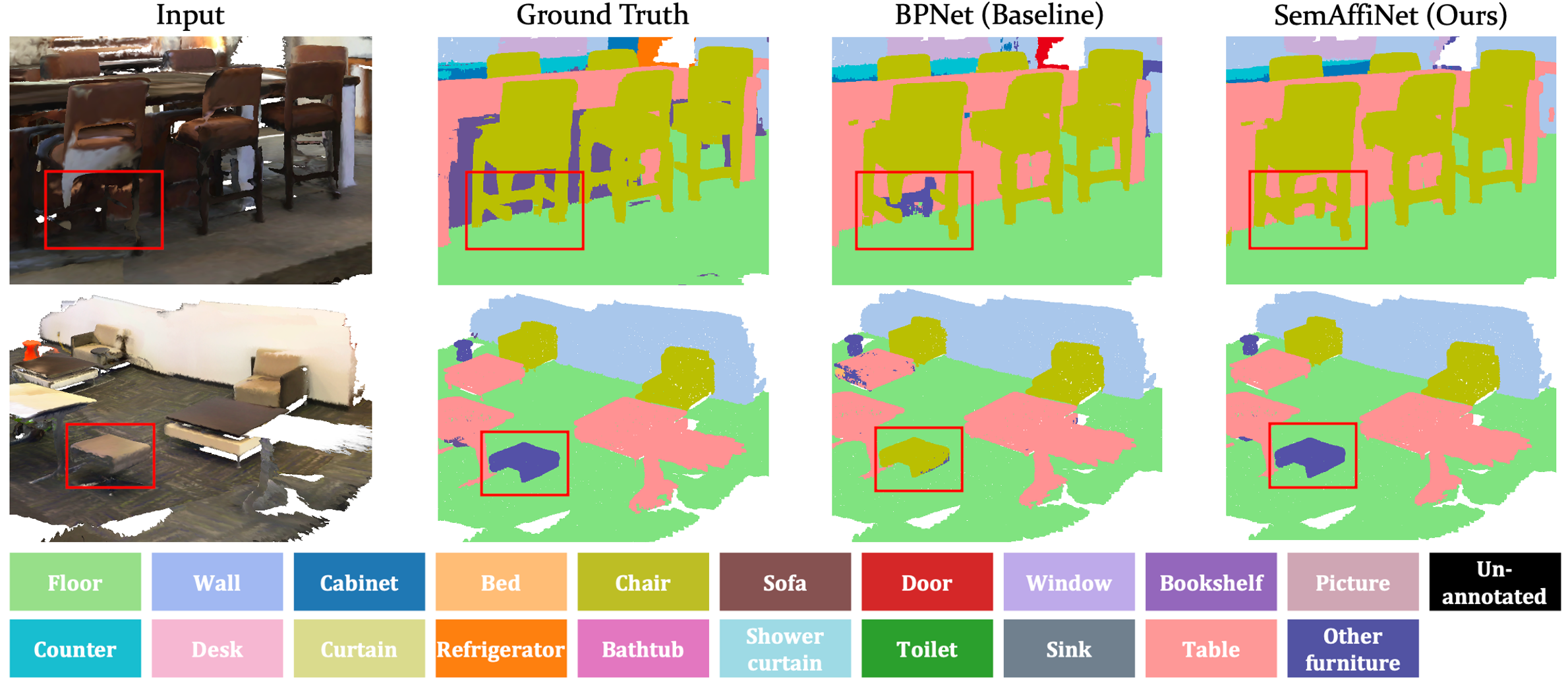}
	\caption{Qualitative results of the SemAffiNet on ScanNetV2 point cloud semantic segmentation task. The first and second columns are the input point clouds and corresponding ground truth labels. The third and fourth columns are the segmentation results of BPNet baseline and our method SemAffiNet respectively. As is shown in the red rectangle in the first line, SemAffiNet is able to recognize \textit{chair crossbeam} while BPNet cannot. From the second line, SemAffiNet correctly identifies that the object in the red rectangle is \textit{not chair}, while BPNet fails. The final line shows the correspondence between categories and visualization colors.}
	\label{fig:results}
	\vspace{-3mm}
\end{figure*}

\subsection{Main Results}
\label{sec:main_result}
Following BPNet~\cite{hu2021bidirectional}, we conduct semantic segmentation experiments on ScanNetV2, evaluating  both 2cm and 5cm voxel settings on validation set. 
The quantitative results are shown in Table~\ref{tab:san}.
Under 5cm settings, SemAffiNet exceeds BPNet baseline by 1.5\% and 3.1\% respectively on 3D mIoU and 2D mIoU metrics. Under 2cm settings, SemAffiNet outperforms BPNet by 2.0\% and 1.5\% respectively on 3D and 2D segmentation results.
We also surpass other previous 3D semantic segmentation methods that take point clouds as input~\cite{qi2017pointnet++, wu2019pointconv,  yan2020pointasnl, thomas2019kpconv, graham20183d, gong2021omni, choy20194d, nekrasov2021mix3d} or use point clouds and auxiliary 2D images as input~\cite{jaritz2019multi, zhang2020fusion}.

Besides ScanNetV2 which mainly focuses on 3D point cloud segmentation, we also conduct experiments on the NYUv2~\cite{NYU2012Silberman} dataset that consists of RGB images and corresponding depth maps. We convert the depth image to pseudo point clouds according to camera pose and employ SemAffiNet. Following BPNet~\cite{hu2021bidirectional}, we adopt the 13-class configuration and report the dense pixel classification accuracy. The experimental results are shown in Table~\ref{tab:nyu}, and our SemAffiNet outperforms those baselines by a large margin, which verifies its superiority.

The quantitative results are shown in Figure~\ref{fig:results}. From the first line example, our SemAffiNet is able to recognize the subtle \textit{chair crossbeam} that BPNet baseline fails to correctly segment. From the second line example, our SemAffiNet correctly classifies the \textit{other furniture} object which is visually similar to \textit{chair}. From these qualitative results, we can prove the superiority of the SemAffiNet over the BPNet baseline. On the one hand, it has the ability to recognize subtle local parts that are easy to be confused with the background. On the other hand, it can correctly classify objects that are visually similar to other categories. 

\begin{table}
  \centering
  \setlength{\abovecaptionskip}{0.2cm}
  \caption{Quantitative results of inserting ESAM into various semantic segmentation baselines. The first two lines show results based on 3D voxel-based MinkowskiNet on the ScanNetV2 dataset. The third line shows results based on 3D point-based KPConv on the S3DIS dataset. The last two lines show results based on SemanticFPN on the 2D Cityscapes dataset.}
  \setlength{\tabcolsep}{3.5mm}
  \begin{tabular}{@{}lcc@{}}
    \toprule
        Method          &   Dataset         &   mIoU(\%) \\
    \midrule
        MinkowskiNet (5cm)\cite{choy20194d} &   ScanNetV2 &   67.4 \\
        + ESAM       &   ScanNetV2       &   \textbf{68.8} \\
    \midrule
        MinkowskiNet (2cm)    &   ScanNetV2 &   72.2 \\
        + ESAM       &   ScanNetV2       &   \textbf{74.0} \\
    \midrule
        KPConv~\cite{thomas2019kpconv}       &   S3DIS  &   65.8 \\
        + ESAM       &   S3DIS  &   \textbf{66.7} \\
    \midrule
        SemanticFPN (Res50)~\cite{kirillov2019panoptic} &   Cityscapes      &   76.1 \\
        + ESAM          &   Cityscapes      &   \textbf{77.2} \\
    \midrule
        SemanticFPN (Res101)    &   Cityscapes  &   77.4 \\
        + ESAM          &   Cityscapes          &   \textbf{79.0} \\
    \bottomrule
  \end{tabular}
  \label{tab:sat}
\end{table}

\begin{table*}[t]
    \centering
    \caption{Ablation studies on individual contributions of SemAffiNet designs on the ScanNetV2 validation dataset under 5cm voxelization.}
    \setlength{\tabcolsep}{1.5mm}
    \begin{tabular}{l|c|c|c|cccc|ccc|c|cc}
    \toprule
    & Ablation on $\downarrow$ & idx
    
    & VN & $\mathrm{\hat{F}}$.FC & $\mathrm{\hat{F}}$.M 
         & $\mathrm{\hat{M}}$.FC & $\mathrm{\hat{M}}$.M 
         & BN & AdaIN & SA & Fuse & 2D mIoU & 3D mIoU \\
    
    \hline
    \multirow{2}{*}{Backbone} & \multirow{2}{*}{View Num}
    & $a$ & \threeg & \cmarkg & \xmarkg & \xmarkg & \xmarkg & \cmarkg & \xmarkg & \xmarkg & \xmarkg & 65.1 & 70.6 \\
    & & $b$ & \five & \cmarkg & \xmarkg & \xmarkg & \xmarkg & \cmarkg & \xmarkg & \xmarkg & \xmarkg & 65.5 & 70.1 \\
    \hline
    \multirow{5}{*}{ESAM} &
    \multirow{3}{*}{\rotatebox{0}{\makecell[c]{Multi-level \\ Segmentation}}}
            & $c$ & \fiveg & \xmark & \cmark & \xmarkg & \xmarkg & \cmarkg & \xmarkg & \xmarkg & \xmarkg & 65.8 & 70.3 \\ 
            & & $d$ & \fiveg & \xmarkg & \cmarkg & \cmark & \xmarkg & \cmarkg & \xmarkg & \xmarkg & \xmarkg & 65.4 & 70.6 \\
            & & $e$ & \fiveg & \xmarkg & \cmarkg & \xmark & \cmark & \cmarkg & \xmarkg & \xmarkg & \xmarkg & 67.2 & 70.8 \\ 
    \cline{2-14}
    & \multirow{2}{*}{\rotatebox{0}{SemAffine}}
            & $f$ & \fiveg & \xmarkg & \cmarkg & \xmarkg & \cmarkg & \xmark & \cmark & \xmarkg & \xmarkg & 66.8 & 71.2 \\
            & & $g$ & \fiveg & \xmarkg & \cmarkg & \xmarkg & \cmarkg & \xmarkg & \xmark & \cmark & \xmarkg & 68.0 & 71.8 \\
    \hline
    ISAM &
    Multi-Modality & $h$ & \fiveg & \xmarkg & \cmarkg & \xmarkg & \cmarkg & \xmarkg & \xmarkg & \cmarkg & \cmark & 68.2 & 72.1 \\
    \hline
    SemAffiNet &
    View Num & $i$ & \three & \xmarkg & \cmarkg & \xmarkg & \cmarkg & \xmarkg & \xmarkg & \cmarkg & \cmarkg & 68.3 & 70.7 \\
    \bottomrule
    \end{tabular}
    \label{tab:abl}
\end{table*}

\subsection{Effects of Semantic-Affine Transformation}
\label{sec:sat_analysis}
Besides SemAffiNet architecture, we also evaluate the semantic-affine transformation independently by wrapping it as a plug-and-play module ESAM and inserting it into various 3D point cloud and 2D image encoder-decoder segmentation baselines. Point cloud methods can be further divided into voxel-based and point-based ones.

For voxel-based 3D segmentation methods, we select MinkowskiNet~\cite{choy20194d} as the baseline given its efficiency and high performance. We conduct semantic segmentation on the ScanNetV2~\cite{dai2017scannet} validation set under both 5cm and 2cm voxelization settings, and the results are shown in the first two lines of Table~\ref{tab:sat}. By inserting ESAM into MinkowskiNet, we reach a higher performance on 3D mIoU results.

For point-based 3D segmentation methods, we choose KPConv~\cite{thomas2019kpconv} (rigid) as the baseline, since it is one of the most classical point-based approaches and has been analyzed as the baseline in many recently published papers~\cite{nekrasov2021mix3d, gong2021omni}. We conduct semantic segmentation on the S3DIS~\cite{armeni20163d} dataset and the results are shown in the third line of Table~\ref{tab:sat}. The qualitative results demonstrate the improvement made by ESAM, convincing the generalization ability of the semantic-affine transformation on both voxel-based and point-based point cloud segmentation methods.

For 2D image segmentation methods, we choose the classical SemanticFPN~\cite{kirillov2019panoptic} as the baseline, implementing ResNet-50 and ResNet-101 settings. We conduct semantic segmentation on the Cityscapes~\cite{Cordts2016Cityscapes} dataset and the results are shown in the last two lines of Table~\ref{tab:sat}. The quantitative results show that the proposed ESAM is not restricted within 3D domain and brings consistent improvements to 2D segmentation baseline under different settings.

\subsection{Ablation Studies}
\label{sec:ablation}
In order to measure the contribution of each SemAffiNet module, we conduct ablation studies on the ScanNetV2 validation set under 5cm setting. ESAM can be divided into two parts: multi-level segmentation and semantic-affine transformation (SA). For multi-level segmentation, we incrementally add mid-level segmentation ($\mathrm{\hat{M}}.$) besides final-level segmentation ($\mathrm{\hat{F}}.$) and replace fully connected classifier (FC) with mask classifier (M). ISAM module fuses multi-modality information (Fuse). The ablation results are shown in Table~\ref{tab:abl} and we conclude that each block makes its own contribution to the overall progression, where semantic-affine transformation is the most effective one. 

Besides ablations on sub-modules of SemAffiNet, we further explore the superiority of the semantic-affine transformation by replacing it with Adaptive Instance Normalization(AdaIN)~\cite{huang2017arbitrary}, whose affine parameters are entirely learned with gradient descent and lack explicit semantic guidance. Comparing line $e,f,g$ in Table~\ref{tab:abl}, AdaIN (line $f$) brings little improvement compared with the vanilla Batch Normalization (line $e$). However, semantic-affine transformation (line $g$) yields more significant progression. 

Furthermore, we also conduct ablation experiments on the number of 2D views (VN). According to line $a,b$, adding views to BPNet~\cite{hu2021bidirectional} baseline doesn't lead to better results but adds computation burden. The reason might be that the potential of the network is exhausted when processing 3 views. However, according to our experiments (line $h$ and $i$), our SemAffiNet performs better on 3D semantic segmentation when increasing view numbers from 3 to 5. The experimental results show that our SemAffiNet reveals more semantic knowledge and has larger potential. 

\subsection{Limitations}
\label{sec:limit}
Even though the proposed semantic-affine transformation is a general representation learning technique to enhance the semantic cognition ability of point clouds features, migrating and performing it on broader point cloud understanding tasks is a non-trivial problem. Since we need per-point supervision on mid-level points, we need to develop weakly-supervised or unsupervised learning techniques when the per-point annotation is inaccessible.

\section{Conclusions}
In this paper, we propose the semantic-affine transformation to explicitly map mid-level features of the backbone decoder to more semantic-distinct embeddings. Based on this technique, we build a semantic-aware segmentation network SemAffiNet. The ESAM explicitly predicts class masks and regresses semantic-affine parameters via the Transformer decoder with learnable query tokens, while the ISAM fuses multi-modal information via the self-attention mechanism. We conduct experiments with the SemAffiNet on the ScanNetV2 dataset and outperform the previous state-of-the-art BPNet. We also prove the generalization ability of the semantic-affine transformation by wrapping it into a plug-and-play ESAM and evaluating it on both 3D point cloud and 2D image segmentation baselines under various settings. 
We believe that the semantic-affine transformation will advance related works in the community, given its simple implementation and reasonable insights. 

\section*{Acknowledgement}
This work was supported in part by the National Natural Science Foundation of China under Grant 62125603 and Grant U1813218, and in part by a grant from the Beijing Academy of Artificial Intelligence (BAAI).

\newpage
{\small
\bibliographystyle{ieee_fullname}
\bibliography{egbib}

\begin{thebibliography}{10}\itemsep=-1pt

\bibitem{aksoy2020salsanet}
Eren~Erdal Aksoy, Saimir Baci, and Selcuk Cavdar.
\newblock Salsanet: Fast road and vehicle segmentation in lidar point clouds
  for autonomous driving.
\newblock In {\em IEEE Intelligent Vehicles Symposium}, pages 926--932, 2020.

\bibitem{alonso20203d}
Inigo Alonso, Luis Riazuelo, Luis Montesano, and Ana~C Murillo.
\newblock 3d-mininet: Learning a 2d representation from point clouds for fast
  and efficient 3d lidar semantic segmentation.
\newblock {\em IEEE Robotics and Automation Letters}, 5(4):5432--5439, 2020.

\bibitem{armeni20163d}
Iro Armeni, Ozan Sener, Amir~R Zamir, Helen Jiang, Ioannis Brilakis, Martin
  Fischer, and Silvio Savarese.
\newblock 3d semantic parsing of large-scale indoor spaces.
\newblock In {\em CVPR}, pages 1534--1543, 2016.

\bibitem{badrinarayanan2017segnet}
Vijay Badrinarayanan, Alex Kendall, and Roberto Cipolla.
\newblock Segnet: A deep convolutional encoder-decoder architecture for image
  segmentation.
\newblock {\em TPAMI}, 39(12):2481--2495, 2017.

\bibitem{carion2020end}
Nicolas Carion, Francisco Massa, Gabriel Synnaeve, Nicolas Usunier, Alexander
  Kirillov, and Sergey Zagoruyko.
\newblock End-to-end object detection with transformers.
\newblock In {\em ECCV}, pages 213--229, 2020.

\bibitem{chen2017deeplab}
Liang-Chieh Chen, George Papandreou, Iasonas Kokkinos, Kevin Murphy, and Alan~L
  Yuille.
\newblock Deeplab: Semantic image segmentation with deep convolutional nets,
  atrous convolution, and fully connected crfs.
\newblock {\em TPAMI}, 40(4):834--848, 2017.

\bibitem{chen2017rethinking}
Liang-Chieh Chen, George Papandreou, Florian Schroff, and Hartwig Adam.
\newblock Rethinking atrous convolution for semantic image segmentation.
\newblock {\em arXiv preprint arXiv:1706.05587}, 2017.

\bibitem{chen2020pointmixup}
Yunlu Chen, Vincent~Tao Hu, Efstratios Gavves, Thomas Mensink, Pascal Mettes,
  Pengwan Yang, and Cees~GM Snoek.
\newblock Pointmixup: Augmentation for point clouds.
\newblock In {\em ECCV}, pages 330--345, 2020.

\bibitem{cheng2021maskformer}
Bowen Cheng, Alexander~G. Schwing, and Alexander Kirillov.
\newblock Per-pixel classification is not all you need for semantic
  segmentation.
\newblock {\em arXiv}, 2021.

\bibitem{cheng20212}
Ran Cheng, Ryan Razani, Ehsan Taghavi, Enxu Li, and Bingbing Liu.
\newblock 2-s3net: Attentive feature fusion with adaptive feature selection for
  sparse semantic segmentation network.
\newblock In {\em CVPR}, pages 12547--12556, 2021.

\bibitem{choy20194d}
Christopher Choy, JunYoung Gwak, and Silvio Savarese.
\newblock 4d spatio-temporal convnets: Minkowski convolutional neural networks.
\newblock In {\em CVPR}, pages 3075--3084, 2019.

\bibitem{Cordts2016Cityscapes}
Marius Cordts, Mohamed Omran, Sebastian Ramos, Timo Rehfeld, Markus Enzweiler,
  Rodrigo Benenson, Uwe Franke, Stefan Roth, and Bernt Schiele.
\newblock The cityscapes dataset for semantic urban scene understanding.
\newblock In {\em CVPR}, 2016.

\bibitem{cortinhal2020salsanext}
Tiago Cortinhal, George Tzelepis, and Eren~Erdal Aksoy.
\newblock Salsanext: Fast, uncertainty-aware semantic segmentation of lidar
  point clouds.
\newblock In {\em ISVC}, pages 207--222, 2020.

\bibitem{dai2017scannet}
Angela Dai, Angel~X Chang, Manolis Savva, Maciej Halber, Thomas Funkhouser, and
  Matthias Nie{\ss}ner.
\newblock Scannet: Richly-annotated 3d reconstructions of indoor scenes.
\newblock In {\em CVPR}, pages 5828--5839, 2017.

\bibitem{dai20183dmv}
Angela Dai and Matthias Nie{\ss}ner.
\newblock 3dmv: Joint 3d-multi-view prediction for 3d semantic scene
  segmentation.
\newblock In {\em ECCV}, pages 452--468, 2018.

\bibitem{deng2021ga}
Shuang Deng and Qiulei Dong.
\newblock Ga-net: Global attention network for point cloud semantic
  segmentation.
\newblock {\em IEEE Signal Processing Letters}, 2021.

\bibitem{dosovitskiy2020image}
Alexey Dosovitskiy, Lucas Beyer, Alexander Kolesnikov, Dirk Weissenborn,
  Xiaohua Zhai, Thomas Unterthiner, Mostafa Dehghani, Matthias Minderer, Georg
  Heigold, Sylvain Gelly, et~al.
\newblock An image is worth 16x16 words: Transformers for image recognition at
  scale.
\newblock {\em arXiv preprint arXiv:2010.11929}, 2020.

\bibitem{fan2021scf}
Siqi Fan, Qiulei Dong, Fenghua Zhu, Yisheng Lv, Peijun Ye, and Fei-Yue Wang.
\newblock Scf-net: Learning spatial contextual features for large-scale point
  cloud segmentation.
\newblock In {\em CVPR}, pages 14504--14513, 2021.

\bibitem{gong2021omni}
Jingyu Gong, Jiachen Xu, Xin Tan, Haichuan Song, Yanyun Qu, Yuan Xie, and
  Lizhuang Ma.
\newblock Omni-supervised point cloud segmentation via gradual receptive field
  component reasoning.
\newblock In {\em CVPR}, pages 11673--11682, 2021.

\bibitem{graham20183d}
Benjamin Graham, Martin Engelcke, and Laurens Van Der~Maaten.
\newblock 3d semantic segmentation with submanifold sparse convolutional
  networks.
\newblock In {\em CVPR}, pages 9224--9232, 2018.

\bibitem{guo2021pct}
Meng-Hao Guo, Jun-Xiong Cai, Zheng-Ning Liu, Tai-Jiang Mu, Ralph~R Martin, and
  Shi-Min Hu.
\newblock Pct: Point cloud transformer.
\newblock {\em Computational Visual Media}, 7(2):187--199, 2021.

\bibitem{guo2021sotr}
Ruohao Guo, Dantong Niu, Liao Qu, and Zhenbo Li.
\newblock Sotr: Segmenting objects with transformers.
\newblock In {\em ICCV}, pages 7157--7166, 2021.

\bibitem{handa2016understanding}
Ankur Handa, Viorica Patraucean, Vijay Badrinarayanan, Simon Stent, and Roberto
  Cipolla.
\newblock Understanding real world indoor scenes with synthetic data.
\newblock In {\em CVPR}, pages 4077--4085, 2016.

\bibitem{he2016deep}
Kaiming He, Xiangyu Zhang, Shaoqing Ren, and Jian Sun.
\newblock Deep residual learning for image recognition.
\newblock In {\em CVPR}, pages 770--778, 2016.

\bibitem{hermans2014dense}
Alexander Hermans, Georgios Floros, and Bastian Leibe.
\newblock Dense 3d semantic mapping of indoor scenes from rgb-d images.
\newblock In {\em ICRA}, pages 2631--2638, 2014.

\bibitem{hu2020randla}
Qingyong Hu, Bo Yang, Linhai Xie, Stefano Rosa, Yulan Guo, Zhihua Wang, Niki
  Trigoni, and Andrew Markham.
\newblock Randla-net: Efficient semantic segmentation of large-scale point
  clouds.
\newblock In {\em CVPR}, pages 11108--11117, 2020.

\bibitem{hu2021bidirectional}
Wenbo Hu, Hengshuang Zhao, Li Jiang, Jiaya Jia, and Tien-Tsin Wong.
\newblock Bidirectional projection network for cross dimension scene
  understanding.
\newblock In {\em CVPR}, pages 14373--14382, 2021.

\bibitem{huang2019texturenet}
Jingwei Huang, Haotian Zhang, Li Yi, Thomas Funkhouser, Matthias Nie{\ss}ner,
  and Leonidas~J Guibas.
\newblock Texturenet: Consistent local parametrizations for learning from
  high-resolution signals on meshes.
\newblock In {\em CVPR}, pages 4440--4449, 2019.

\bibitem{huang2017arbitrary}
Xun Huang and Serge Belongie.
\newblock Arbitrary style transfer in real-time with adaptive instance
  normalization.
\newblock In {\em ICCV}, pages 1501--1510, 2017.

\bibitem{jaritz2019multi}
Maximilian Jaritz, Jiayuan Gu, and Hao Su.
\newblock Multi-view pointnet for 3d scene understanding.
\newblock In {\em ICCV Workshops}, pages 0--0, 2019.

\bibitem{ke2018multi}
Lipeng Ke, Ming-Ching Chang, Honggang Qi, and Siwei Lyu.
\newblock Multi-scale structure-aware network for human pose estimation.
\newblock In {\em ECCV}, pages 713--728, 2018.

\bibitem{kirillov2019panoptic}
Alexander Kirillov, Ross Girshick, Kaiming He, and Piotr Doll{\'a}r.
\newblock Panoptic feature pyramid networks.
\newblock In {\em CVPR}, pages 6399--6408, 2019.

\bibitem{klokov2017escape}
Roman Klokov and Victor Lempitsky.
\newblock Escape from cells: Deep kd-networks for the recognition of 3d point
  cloud models.
\newblock In {\em ICCV}, pages 863--872, 2017.

\bibitem{li2018adaptive}
Ruoyu Li, Sheng Wang, Feiyun Zhu, and Junzhou Huang.
\newblock Adaptive graph convolutional neural networks.
\newblock In {\em AAAI}, volume~32, 2018.

\bibitem{li2018pointcnn}
Yangyan Li, Rui Bu, Mingchao Sun, Wei Wu, Xinhan Di, and Baoquan Chen.
\newblock Pointcnn: Convolution on x-transformed points.
\newblock {\em NeurIPS}, 31:820--830, 2018.

\bibitem{liu2021exploit}
Mingyuan Liu, Dan Schonfeld, and Wei Tang.
\newblock Exploit visual dependency relations for semantic segmentation.
\newblock In {\em CVPR}, pages 9726--9735, 2021.

\bibitem{liu2021swin}
Ze Liu, Yutong Lin, Yue Cao, Han Hu, Yixuan Wei, Zheng Zhang, Stephen Lin, and
  Baining Guo.
\newblock Swin transformer: Hierarchical vision transformer using shifted
  windows.
\newblock {\em arXiv preprint arXiv:2103.14030}, 2021.

\bibitem{liu2019point}
Zhijian Liu, Haotian Tang, Yujun Lin, and Song Han.
\newblock Point-voxel cnn for efficient 3d deep learning.
\newblock {\em arXiv preprint arXiv:1907.03739}, 2019.

\bibitem{long2015fully}
Jonathan Long, Evan Shelhamer, and Trevor Darrell.
\newblock Fully convolutional networks for semantic segmentation.
\newblock In {\em CVPR}, pages 3431--3440, 2015.

\bibitem{lu2021cga}
Tao Lu, Limin Wang, and Gangshan Wu.
\newblock Cga-net: Category guided aggregation for point cloud semantic
  segmentation.
\newblock In {\em CVPR}, pages 11693--11702, 2021.

\bibitem{maturana2015voxnet}
Daniel Maturana and Sebastian Scherer.
\newblock Voxnet: A 3d convolutional neural network for real-time object
  recognition.
\newblock In {\em IROS}, pages 922--928, 2015.

\bibitem{mccormac2017semanticfusion}
John McCormac, Ankur Handa, Andrew Davison, and Stefan Leutenegger.
\newblock Semanticfusion: Dense 3d semantic mapping with convolutional neural
  networks.
\newblock In {\em ICRA}, pages 4628--4635, 2017.

\bibitem{milioto2019rangenet++}
Andres Milioto, Ignacio Vizzo, Jens Behley, and Cyrill Stachniss.
\newblock Rangenet++: Fast and accurate lidar semantic segmentation.
\newblock In {\em IROS}, pages 4213--4220, 2019.

\bibitem{NYU2012Silberman}
Pushmeet~Kohli Nathan~Silberman, Derek~Hoiem and Rob Fergus.
\newblock Indoor segmentation and support inference from rgbd images.
\newblock In {\em ECCV}, 2012.

\bibitem{nekrasov2021mix3d}
Alexey Nekrasov, Jonas Schult, Or Litany, Bastian Leibe, and Francis Engelmann.
\newblock Mix3d: Out-of-context data augmentation for 3d scenes.
\newblock {\em arXiv preprint arXiv:2110.02210}, 2021.

\bibitem{paszke2019pytorch}
Adam Paszke, Sam Gross, Francisco Massa, Adam Lerer, James Bradbury, Gregory
  Chanan, Trevor Killeen, Zeming Lin, Natalia Gimelshein, Luca Antiga, et~al.
\newblock Pytorch: An imperative style, high-performance deep learning library.
\newblock {\em NeurIPS}, 32:8026--8037, 2019.

\bibitem{qi2017pointnet}
Charles~R Qi, Hao Su, Kaichun Mo, and Leonidas~J Guibas.
\newblock Pointnet: Deep learning on point sets for 3d classification and
  segmentation.
\newblock In {\em CVPR}, pages 652--660, 2017.

\bibitem{qi2017pointnet++}
Charles~R Qi, Li Yi, Hao Su, and Leonidas~J Guibas.
\newblock Pointnet++: Deep hierarchical feature learning on point sets in a
  metric space.
\newblock {\em arXiv preprint arXiv:1706.02413}, 2017.

\bibitem{qiu2021semantic}
Shi Qiu, Saeed Anwar, and Nick Barnes.
\newblock Semantic segmentation for real point cloud scenes via bilateral
  augmentation and adaptive fusion.
\newblock In {\em CVPR}, pages 1757--1767, 2021.

\bibitem{quan2016fusionnet}
Tran~Minh Quan, David~GC Hildebrand, and Won-Ki Jeong.
\newblock Fusionnet: A deep fully residual convolutional neural network for
  image segmentation in connectomics.
\newblock {\em arXiv preprint arXiv:1612.05360}, 2016.

\bibitem{rethage2018fully}
Dario Rethage, Johanna Wald, Jurgen Sturm, Nassir Navab, and Federico Tombari.
\newblock Fully-convolutional point networks for large-scale point clouds.
\newblock In {\em ECCV}, pages 596--611, 2018.

\bibitem{riegler2017octnet}
Gernot Riegler, Ali Osman~Ulusoy, and Andreas Geiger.
\newblock Octnet: Learning deep 3d representations at high resolutions.
\newblock In {\em CVPR}, pages 3577--3586, 2017.

\bibitem{ronneberger2015u}
Olaf Ronneberger, Philipp Fischer, and Thomas Brox.
\newblock U-net: Convolutional networks for biomedical image segmentation.
\newblock In {\em MICCAI}, pages 234--241, 2015.

\bibitem{ruder2016overview}
Sebastian Ruder.
\newblock An overview of gradient descent optimization algorithms.
\newblock {\em arXiv preprint arXiv:1609.04747}, 2016.

\bibitem{su2018splatnet}
Hang Su, Varun Jampani, Deqing Sun, Subhransu Maji, Evangelos Kalogerakis,
  Ming-Hsuan Yang, and Jan Kautz.
\newblock Splatnet: Sparse lattice networks for point cloud processing.
\newblock In {\em CVPR}, pages 2530--2539, 2018.

\bibitem{tang2020searching}
Haotian Tang, Zhijian Liu, Shengyu Zhao, Yujun Lin, Ji Lin, Hanrui Wang, and
  Song Han.
\newblock Searching efficient 3d architectures with sparse point-voxel
  convolution.
\newblock In {\em ECCV}, pages 685--702, 2020.

\bibitem{thomas2019kpconv}
Hugues Thomas, Charles~R Qi, Jean-Emmanuel Deschaud, Beatriz Marcotegui,
  Fran{\c{c}}ois Goulette, and Leonidas~J Guibas.
\newblock Kpconv: Flexible and deformable convolution for point clouds.
\newblock In {\em ICCV}, pages 6411--6420, 2019.

\bibitem{vaswani2017attention}
Ashish Vaswani, Noam Shazeer, Niki Parmar, Jakob Uszkoreit, Llion Jones,
  Aidan~N Gomez, {\L}ukasz Kaiser, and Illia Polosukhin.
\newblock Attention is all you need.
\newblock In {\em NeurIPS}, pages 5998--6008, 2017.

\bibitem{wang2019dynamic}
Yue Wang, Yongbin Sun, Ziwei Liu, Sanjay~E Sarma, Michael~M Bronstein, and
  Justin~M Solomon.
\newblock Dynamic graph cnn for learning on point clouds.
\newblock {\em TOG}, 38(5):1--12, 2019.

\bibitem{wei2016convolutional}
Shih-En Wei, Varun Ramakrishna, Takeo Kanade, and Yaser Sheikh.
\newblock Convolutional pose machines.
\newblock In {\em CVPR}, pages 4724--4732, 2016.

\bibitem{wu2018squeezeseg}
Bichen Wu, Alvin Wan, Xiangyu Yue, and Kurt Keutzer.
\newblock Squeezeseg: Convolutional neural nets with recurrent crf for
  real-time road-object segmentation from 3d lidar point cloud.
\newblock In {\em ICRA}, pages 1887--1893, 2018.

\bibitem{wu2019squeezesegv2}
Bichen Wu, Xuanyu Zhou, Sicheng Zhao, Xiangyu Yue, and Kurt Keutzer.
\newblock Squeezesegv2: Improved model structure and unsupervised domain
  adaptation for road-object segmentation from a lidar point cloud.
\newblock In {\em ICRA}, pages 4376--4382, 2019.

\bibitem{wu2019pointconv}
Wenxuan Wu, Zhongang Qi, and Li Fuxin.
\newblock Pointconv: Deep convolutional networks on 3d point clouds.
\newblock In {\em CVPR}, pages 9621--9630, 2019.

\bibitem{xu2021paconv}
Mutian Xu, Runyu Ding, Hengshuang Zhao, and Xiaojuan Qi.
\newblock Paconv: Position adaptive convolution with dynamic kernel assembling
  on point clouds.
\newblock In {\em CVPR}, pages 3173--3182, 2021.

\bibitem{yan2020pointasnl}
Xu Yan, Chaoda Zheng, Zhen Li, Sheng Wang, and Shuguang Cui.
\newblock Pointasnl: Robust point clouds processing using nonlocal neural
  networks with adaptive sampling.
\newblock In {\em CVPR}, pages 5589--5598, 2020.

\bibitem{ye2021drinet}
Maosheng Ye, Shuangjie Xu, Tongyi Cao, and Qifeng Chen.
\newblock Drinet: A dual-representation iterative learning network for point
  cloud segmentation.
\newblock In {\em ICCV}, pages 7447--7456, 2021.

\bibitem{yu2021pointr}
Xumin Yu, Yongming Rao, Ziyi Wang, Zuyan Liu, Jiwen Lu, and Jie Zhou.
\newblock Pointr: Diverse point cloud completion with geometry-aware
  transformers.
\newblock In {\em ICCV}, pages 12498--12507, 2021.

\bibitem{zhang2021pointcutmix}
Jinlai Zhang, Lyujie Chen, Bo Ouyang, Binbin Liu, Jihong Zhu, Yujing Chen,
  Yanmei Meng, and Danfeng Wu.
\newblock Pointcutmix: Regularization strategy for point cloud classification.
\newblock {\em arXiv preprint arXiv:2101.01461}, 2021.

\bibitem{zhang2020fusion}
Jiazhao Zhang, Chenyang Zhu, Lintao Zheng, and Kai Xu.
\newblock Fusion-aware point convolution for online semantic 3d scene
  segmentation.
\newblock In {\em CVPR}, pages 4534--4543, 2020.

\bibitem{zhang2020polarnet}
Yang Zhang, Zixiang Zhou, Philip David, Xiangyu Yue, Zerong Xi, Boqing Gong,
  and Hassan Foroosh.
\newblock Polarnet: An improved grid representation for online lidar point
  clouds semantic segmentation.
\newblock In {\em CVPR}, pages 9601--9610, 2020.

\bibitem{zhao2021point}
Hengshuang Zhao, Li Jiang, Jiaya Jia, Philip~HS Torr, and Vladlen Koltun.
\newblock Point transformer.
\newblock In {\em ICCV}, pages 16259--16268, 2021.

\bibitem{zhao2017pyramid}
Hengshuang Zhao, Jianping Shi, Xiaojuan Qi, Xiaogang Wang, and Jiaya Jia.
\newblock Pyramid scene parsing network.
\newblock In {\em CVPR}, pages 2881--2890, 2017.

\bibitem{zheng2021rethinking}
Sixiao Zheng, Jiachen Lu, Hengshuang Zhao, Xiatian Zhu, Zekun Luo, Yabiao Wang,
  Yanwei Fu, Jianfeng Feng, Tao Xiang, Philip~HS Torr, et~al.
\newblock Rethinking semantic segmentation from a sequence-to-sequence
  perspective with transformers.
\newblock In {\em CVPR}, pages 6881--6890, 2021.

\bibitem{zhou2020cylinder3d}
Hui Zhou, Xinge Zhu, Xiao Song, Yuexin Ma, Zhe Wang, Hongsheng Li, and Dahua
  Lin.
\newblock Cylinder3d: An effective 3d framework for driving-scene lidar
  semantic segmentation.
\newblock {\em arXiv preprint arXiv:2008.01550}, 2020.

\end{thebibliography}
}

\newpage
\clearpage
\begin{appendix}

In this supplementary material, we provide the detailed network architecture of SemAffiNet in Section~\ref{sec:arch}. Then in Section~\ref{sec:exp}, we show additional experiments information, including the datasets introduction, implementation details and class-wise 3D point cloud segmentation mIoU on the ScanNetV2~\cite{dai2017scannet} dataset.

\section{Network Architecture}
\label{sec:arch}

\subsection{The Backbone}
We choose the BPNet~\cite{hu2021bidirectional} as our backbone, which consists of a 3D sparse convolution encoder-decoder branch and a 2D vanilla convolution encoder-decoder branch. The 3D branch implements the sparse convolution MinkowskiEngine~\cite{choy20194d} to build a ResNet-style architecture~\cite{he2016deep}, while the 2D branch exploits the ResNet34 model as the encoder. The decoder blocks of these two branches are composed of residual blocks with interpolation operation for the 2D branch and transpose convolution for the 3D branch. Skip connections are added between the corresponding layers of encoders and decoders to pass low-level information. The bidirectional links between layers of the 2D branch and the 3D branch from the BPNet are kept to communicate knowledge between the two modalities. Please refer to the BPNet paper for further details about the bidirectional projections between 2D and 3D features.

\subsection{The Implicit Semantic-Aware Module}
The ISAM is a shared Transformer~\cite{vaswani2017attention} encoder module that consists of $N_e=6$ encoder blocks to fuse 2D and 3D multi-modal information. Each encoder block consists of a Multi-Head Attention layer with $8$ heads, two LayerNorm layers and a Feed Forward layer. The input to the Transformer encoder is the concatenation of the high-level features from both 2D and 3D branches.

\subsection{The Explicit Semantic-Aware Module}
We design two similar explicit semantic-aware modules (ESAM) for 2D and 3D branches respectively. We wrap the learning parameters of ESAM into a Transformer~\cite{vaswani2017attention} decoder module, predicting semantic-affine parameters and semantic class masks. Then semantic-affine transformations are then applied to multiple mid-level layers of the backbone decoder.

The \textbf{Transformer decoder} consists of $N_d=6$ decoder blocks. Each decoder block consists of a Multi-Head Attention layer with $8$ heads, three LayerNorm layers, a Multi-Head Cross Attention layer with $8$ heads and a Feed Forward layer. The input to the Transformer decoder is the output of the ISAM in the corresponding modality and the learnable class queries of shape $(N,d_h)$, where $N$ is the number of classes and $d_h=128$ is the hidden dimension. The position embeddings for the 2D branch are learnable embeddings with $d_h$ dimensions. We obtain the position embeddings for the 3D branch by feeding the coordinates of 3D point clouds to an MLP block with 3 linear layers. 

The output features $h^u$ of shape $(N, d_h)$ from the Transformer decoder layer $u$ are prepared for predicting semantic-affine parameters and the output feature $h^{-1}$ from the final layer of the Transformer decoder are used for predicting semantic class masks. Firstly, the output features of the Transformer decoder $h^{-1}$ are fed through an MLP block with 3 linear layers to get the class masks $M$ of shape $(N, d_m)$, where $d_m=128$ is the class mask dimension. Secondly, the intermediate output features $h^u$ of the Transformer decoder layer $u$ are fed through two separate MLP blocks with 5 linear layers to obtain class-specific affine parameters $\mathbf{s}^i, \mathbf{b}^i$ for the backbone decoder layer $i$, where the correspondence between $i$ and $u$ is $u=i+2$.

The \textbf{Semantic-Affine Transformation} is implemented to $3$ mid-level layers of the backbone decoders. Once obtaining the backbone decoder feature $f^i$ at layer $i$, we can calculate the per-point or per-pixel classification prediction confidence as $A^i=Mf^i$. Then the semantic-affine transformation parameters for each point $S^i, B^i$ can be calculated via Equation(1) in our main paper according to $\mathbf{s}^i, \mathbf{b}^i$ and $A^i$. Then we transform the normalized feature $\hat{f}^i_{j}$ of the backbone decoders with Equation(2) in our main paper to enhance semantic information of mid-level features.

\begin{table*}[!t]
    \scriptsize
    \centering
    \renewcommand{\tabcolsep}{1.25pt}
    \renewcommand{\arraystretch}{1.3}
    \caption{The class-wise segmentation results on the 3D point cloud segmentation task of the ScanNetV2~\cite{dai2017scannet} dataset. We compare the proposed SemAffiNet with the BPNet~\cite{hu2021bidirectional} baseline under both 5cm and 2cm settings.}
    \resizebox{1.0\linewidth}{!}{
        \begin{tabular}{l|c|cccccccccccccccccccc}
            \toprule[1pt] 
            Method              &  mIoU  &  bath  &  bed   & bkshf  &  cab   & chair  &  cntr  &  curt  &  desk  &  door  & floor  & other  &  pic   & fridge & shower &  sink  &  sofa  & table  & toilet &  wall  & window \\ \hline
            BPNet (5cm)         &  70.6  &  85.6  &  81.6  &  79.8  &  68.7  &  89.9  &  66.3  &  60.0  &  69.2  &  58.6  &  94.6  &  58.0  &  20.9  &  54.7  &  64.9  &  68.4  &  79.7  &  76.6  &  91.4  &  82.9  &  59.4  \\
            SemAffiNet (5cm)    &  72.1  &  85.9  &  80.8  &  82.7  &  69.9  &  90.7  &  65.6  &  66.3  &  71.8  &  61.7  &  94.5  &  56.9  &  26.0  &  52.0  &  71.4  &  66.0  &  82.3  &  76.8  &  92.6  &  83.8  &  65.0  \\ \hline
            BPNet (2cm)         &  72.5  &  86.7  &  79.5  &  80.1  &  66.9  &  90.8  &  62.3  &  74.9  &  69.3  &  63.3  &  95.0  &  56.3  &  34.1  &  55.6  &  71.5  &  65.9  &  83.1  &  73.7  &  92.6  &  84.9  &  63.5  \\
            SemAffiNet (2cm)    &  74.5  &  88.5  &  82.1  &  81.6  &  69.9  &  91.6  &  67.2  &  79.2  &  70.0  &  67.3  &  95.3  &  58.0  &  32.6  &  58.3  &  70.9  &  70.5  &  83.1  &  77.4  &  94.5  &  85.9  &  65.3  \\
            \bottomrule[1pt]
    \end{tabular}}
    \label{tab:class-wise}
\end{table*}

\section{Additional Experiments Information}
\label{sec:exp}

\subsection{Datasets Introduction}
We evaluate our SemAffiNet with both point cloud semantic segmentation task on the ScanNetV2 dataset and RGB-D image segmentation task on the RGB-D NYUv2 dataset. We also verify the generalization ability of the proposed semantic-affine transformation on pure 3D S3DIS dataset and pure 2D Cityscapes dataset. In this subsection, we will introduce these datasets in detail.

\vspace{4pt}

\noindent \textbf{ScanNetV2}~\cite{dai2017scannet} is one of the most commonly used indoor scenes datasets. It consists of over 1500 scans of indoor scenes annotated with 20 commonly seen semantic classes. The RGB-D video dataset provides both 2D image-level color data and 3D point-cloud geometry data, thus being robust and comprehensive for 3D scene understanding. We follow the official split to train on 1201 training scans and test on 312 validation scans.

\vspace{4pt}

\noindent \textbf{S3DIS}~\cite{armeni20163d} is another indoor scene dataset. It contains dense 3D point clouds extracted from 6 large-scale areas scanned from 271 rooms in 3 buildings and is annotated by 13 semantic classes. We follow the common protocol to split Area 5 as the test set and use other Areas for training. 

\vspace{4pt}

\noindent \textbf{NYUv2}~\cite{NYU2012Silberman} is a popular RGB-D dataset that focuses on 2D image segmentation. It consists of 1,449 pairs of aligned RGB and depth images and we follow the official split to train on 795 samples and leave 654 for testing. We convert the depth image to pseudo point clouds according to camera pose, and the problem is transformed into 3D-2D multi-modality segmentation as ScanNetV2.

\vspace{4pt}

\noindent \textbf{Cityscapes}~\cite{Cordts2016Cityscapes} is an outdoor 2D street-scene dataset that contains 2975 train images, 500 validation images and 1525 test images. The images have 1024*2048 resolutions and there are 19 classes. We do not use additional 20k images with coarse annotations.

\subsection{Implementation Details}
We implement our proposed SemAffiNet architecture with PyTorch~\cite{paszke2019pytorch} and utilize SGD optimizer~\cite{ruder2016overview} with base learning rate $2e^{-2}$ and weight decay $1e^{-4}$. The learning rate of parameters in ISAM and ESAM is reduced by a factor of 0.1 for more stable training. We implement a squared learning rate scheduler with a warming up process. We train the model for 100 epochs with batch size 16. 
The loss weights for vanilla 3D and 2D segmentation Cross Entropy loss are $1$ and $0.1$ respectively, following BPNet settings. The loss weights for the auxiliary mid-level Binary Cross Entropy loss for 3D and 2D branches are kept at $1$ for our experiments. 

\subsection{Class-wise Segmentation Results}

The class-wise segmentation results on the 3D point cloud segmentation task of ScanNetV2~\cite{dai2017scannet} validation set are shown in Table~\ref{tab:class-wise}. We compare the proposed SemAffiNet with the BPNet baseline under both 5cm and 2cm settings. From the class-wise segmentation results, we can conclude that our SemAffiNet achieves a higher IoU on most categories and a higher mIoU than the BPNet baseline.

\end{appendix}

\end{document}